\documentclass[conference]{IEEEtran}
\IEEEoverridecommandlockouts
\usepackage{cite}
\usepackage{amsmath,amssymb,amsfonts}
\usepackage{algorithmic}
\usepackage{graphicx}
\usepackage{textcomp}
\usepackage{xcolor}
\def\BibTeX{{\rm B\kern-.05em{\sc i\kern-.025em b}\kern-.08em
    T\kern-.1667em\lower.7ex\hbox{E}\kern-.125emX}}
\usepackage{multirow}
\usepackage{array, booktabs}
\newcolumntype{M}[1]{>{\centering\arraybackslash}m{#1}}

\graphicspath{ {images/} }

\begin{document}

\title{Neural Network-Hardware Co-design for Scalable RRAM-based BNN Accelerators
}

\author{
Yulhwa Kim, Hyungjun Kim, and Jae-Joon Kim\\
\small{Dept. of Creative IT Engineering, Pohang University of Science and Technology (POSTECH),}\\
\small{77 Cheongam-Ro, Nam-Gu, Pohang, Gyeongbuk, 37673, Korea}\\ 
\small{E-mail: {yulhwa.kim, hyungjun.kim, and jaejoon}@postech.ac.kr}\\
}


\maketitle

\begin{abstract}
Recently, RRAM-based Binary Neural Network (BNN) hardware has been gaining interests as it requires 1-bit sense-amp only and eliminates the need for high-resolution ADC and DAC. However, RRAM-based BNN hardware still requires high-resolution ADC for partial sum calculation to implement large-scale neural network using multiple memory arrays. We propose a neural network-hardware co-design approach to split input to fit each split network on a RRAM array so that the reconstructed BNNs calculate 1-bit output neuron in each array. As a result, ADC can be completely eliminated from the design even for large-scale neural network. Simulation results show that the proposed network reconstruction and retraining recovers the inference accuracy of the original BNN. The accuracy loss of the proposed scheme in the CIFAR-10 testcase was less than 1.1\% compared to the original network. The code for training and running proposed BNN models is available at: https://github.com/YulhwaKim/RRAMScalable\_BNN.
\end{abstract}


\section{Introduction}
Deep Neural Networks (DNNs) have been receiving attention as a breakthrough technology for solving complex cognitive problems such as image recognition, speech recognition, and machine translation. As the DNNs require large number of vector-matrix multiplication and memory accesses, DNN computing with conventional processors experiences large delay and energy consumption due to heavy communication between main memory and processors. As a result, design of dedicated hardware for DNNs has been actively studied \cite{Eyeriss, ISAAC, PRIME}. Among them, RRAM-based accelerator is gaining interests for its in-memory computing characteristics and parallel vector-matrix multiplication potential \cite{ISAAC, PRIME}.

As RRAM-based accelerators store the synaptic weights on chip and use memory itself as a computing element, they do not need the high-speed communication between main memory and processing elements \cite{ISAAC, PRIME}. In addition, in the RRAM-based accelerators, vector-matrix multiplication can be performed in parallel by applying analog input voltages to multiple word lines and reading the analog bit line current.   However, the analog results typically need to be converted into a digital format, and area and power consumption for the analog-digital interfaces substantially increase as the bit-resolution of the input and output increases \cite{ISAAC, PRIME}. Therefore, the interface overhead for high-resolution analog-to-digital converter (ADC) and digital-to-analog converter (DAC) needs to be reduced to design more efficient accelerators.

One of the promising solutions for reducing analog-to-digital interface overhead is to use Binary Neural Networks (BNNs) \cite{BNN_bengio, XNOR-NET}, which use 1-bit weight and neuron values. As the 1-bit weight and input requirement eliminates the need for ADC and DAC from the RRAM-based BNN accelerator, the BNN accelerator can use 1-bit word-line drivers and sense-amplifiers (SAs) similar to conventional memory \cite{FPRRAM}.

However, the attractive characteristic of RRAM-based BNN accelerator to use 1-bit SA can be used only when the 1-bit binary activation for an output neuron is computed on a memory array. Unfortunately, the number of rows for a memory array is typically limited under 1K and the number of input neurons is often larger than the number of rows of an array. In such cases, an array needs to compute a high-resolution partial sum using an ADC instead of 1-bit output and the partial sums from multiple arrays need to be added to decide final activation value \cite{BCNN-RRAM, XNOR-RRAM}. Under such a scenario, the area and power consumption problem of analog-to-digital interfaces still exist even in RRAM-based BNN accelerators. Previous works tried to reduce the interface burden by quantizing the partial sum values, thereby reducing the ADC bit-resolution to 4 bits or 3 bits \cite{BCNN-RRAM, XNOR-RRAM}. However, they still cannot generate the output using the 1-bit SA, and hence, fail to reach the full potential of the RRAM-based BNN accelerator.  

In this paper, we propose a neural network/hardware co-design approach to design efficient and scalable RRAM-based BNN accelerators. Unlike the conventional methods that use BNN models designed without considering hardware constraints, the proposed method reconstructs and trains BNN models by splitting inputs to fit each split network on a memory array. The BNNs reconstructed with proposed method calculate the 1-bit output neuron in each RRAM array, so each memory array can maintain the original benefit of the RRAM-based BNN accelerator, which is using 1-bit SA for output circuit. The code for training and running proposed BNN models is available at: https://github.com/YulhwaKim/RRAMScalable\_BNN.

The rest of the paper is organized as follows: Section II introduces backgrounds and motivation of the work. Section III proposes neural network/hardware co-design method for scalable RRAM-based BNN accelerator with 1-bit SA.  Section IV compares the inference accuracy and power consumption of the proposed method and previous works. Then, Section V concludes the paper.


\begin{figure}[htbp]
\centerline{\includegraphics[width=0.95\linewidth]{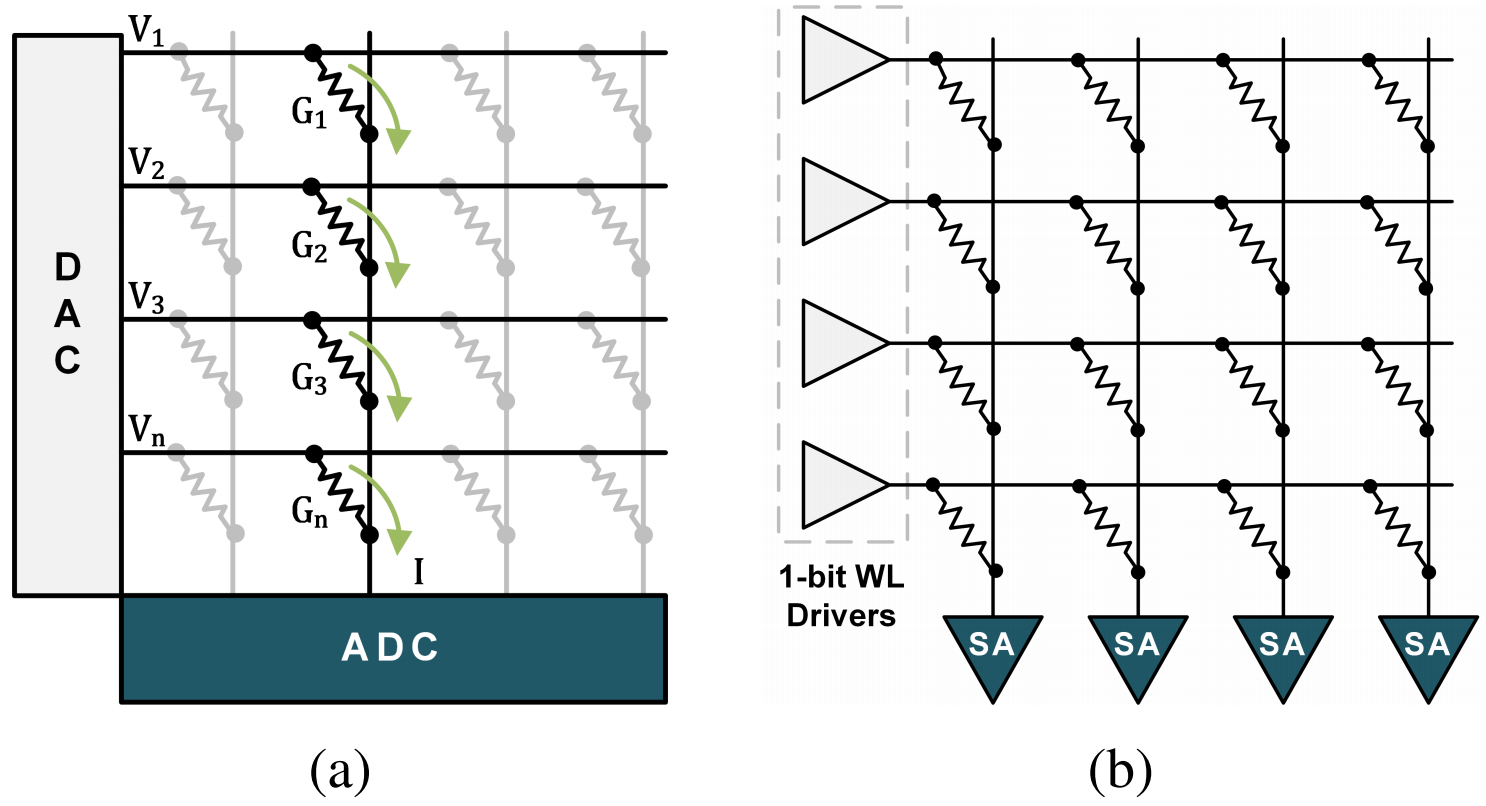}}
\caption{Neural network computational unit based on RRAM array for (a) DNN and (b) BNN. (DAC: Digital-to-analog converters, ADC: Analog-to-digital converters, WL: Word line, SA: Sense amplifier)}
\label{fig:crossbar}
\end{figure}

\begin{figure}[tb]
\centerline{\includegraphics[width=0.95\linewidth]{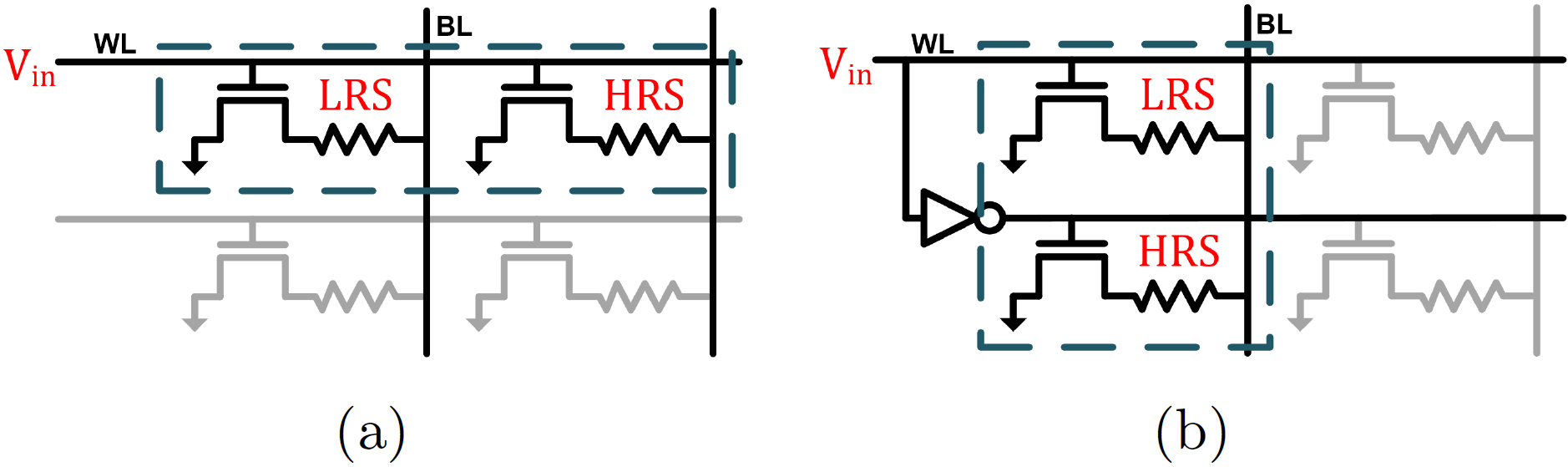}}
\caption{Mapping BNN on a RRAM-based BNN Accelerator when the value of a synaptic weight is +1. Two RRAM cells in a dashed box represent one synaptic weight, and they can represent -1 synaptic weights by exchanging the position of Low Resistance State (LRS) and High Resistance State (HRS).  Each of two implementation methods is widely used for BNN models with (a) (0, 1) neurons and (b) (-1, 1) neurons.}
\label{fig:BNNmapping}
\end{figure}

\begin{figure*}[ht]
\centerline{\includegraphics[width=0.95\linewidth]{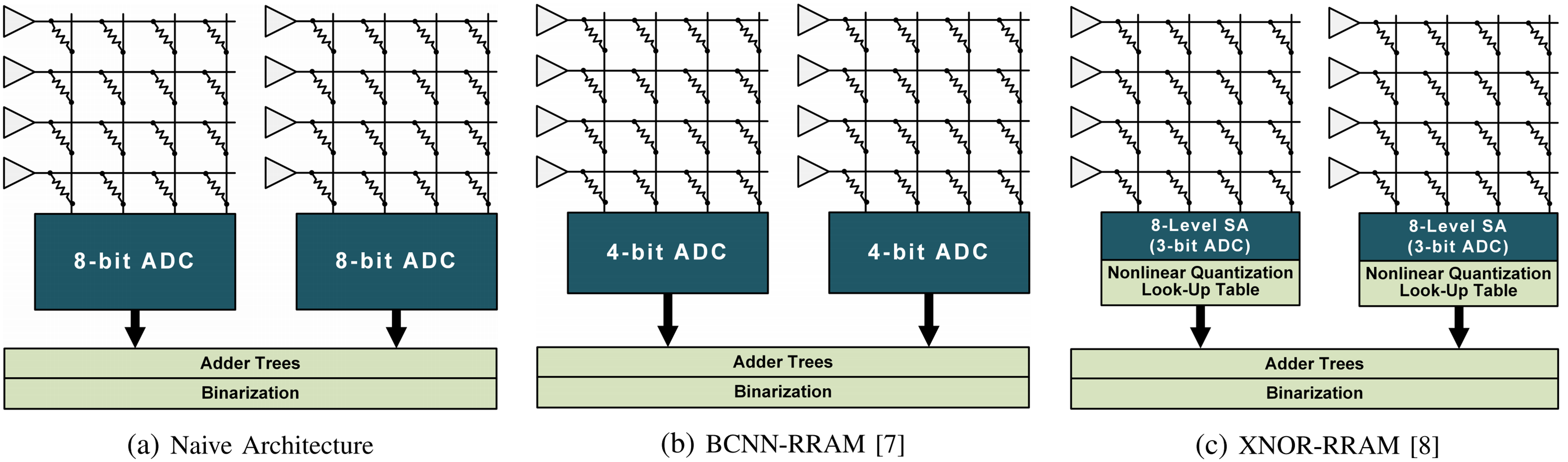}}
\caption{Previous scalable RRAM-based architectures for BNN. (a) represents naive implementation of  BNN. In (b) and (c), partial sums are quantized to reduce the ADC resolution.}
\vspace{-3mm}
\label{fig:architecture_prev}
\end{figure*}

\section{Background}
\subsection{Vector-Matrix Multipliers with RRAM}
RRAM arrays perform vector-matrix multiplication in parallel using analog current. For the operation, the input vector is assigned to the array in the form of input voltages to the word-line drivers of the array, and the 2D weight matrix is assigned in the form of data stored in the 2D RRAM array. Then, according to Ohm's law, the current flowing through each column of the array can be regarded as a multiply-accumulate result as shown in Fig. \ref{fig:crossbar}(a). 

However, to construct a vector-matrix multiplier using the RRAM array, high-resolution ADC and DAC are required to convert the analog results to digital values (Fig. \ref{fig:crossbar}(a)). A critical design issue is that the area and power consumption of the interfaces increase significantly as the resolution of ADC and DAC increases. For instance, ADC/DACs of RRAM-based accelerators consumed more than 73\% of total chip power whereas RRAM array accounted for less than 8.7\% of the total chip power in ISAAC \cite{ISAAC}. In \cite{PRIME}, 1-bit SAs are used as analog-to-digital interfaces, but the SAs are read multiple times to achieve high-resolution output similar to the operating principle of SAR-type ADCs. Thus, the design also suffers from the power issue on the interfaces. Therefore, a new scheme is needed for reducing the interface overhead caused by the high-resolution ADC/DACs.


Recently, binary neural network (BNN), which uses 1-bit weight and 1-bit neuron values, has been gaining attention for its reduced memory footprint and efficient bit-wise computation characteristics \cite{BNN_bengio, XNOR-NET}. The BNN is even more attractive when it is combined with RRAM-based hardware because it uses 1-bit word-line drivers and sense-amplifiers (SAs) instead of high resolution DACs and ADCs as shown in Fig. \ref{fig:crossbar}(b) \cite{FPRRAM}. However, the promising feature can be preserved only when entire calculation for an output neuron value can be done in a memory array. If an array is too small to complete an activation, the array needs to produce a multi-bit partial-sum instead of an 1-bit output. Therefore, the analog-digital interface overhead still remains even in the RRAM-based BNN hardware for large-scale neural network, for which multi-bit partial sums from multiple arrays need to be merged. 

\subsection{Previous Works on Scalable RRAM-based BNN Accelerators}
In the RRAM-based BNN accelerators, the RRAM arrays are used to accelerate multiplication on input neurons and weight matrix of fully-connected layers and convolution layers. For convolution layers, it is common to lower 4D weights to 2D weights of size of (width x height x channel of weight filters) x (number of weight filters). However, because of the physical design constraints of the RRAM array, the number of rows and columns in each array is limited to 1K or less, and the number of input neurons that each crossbar array can process is limited to the number of rows or the half of number of rows depending on BNN models and implementation methods as shown in Fig. \ref{fig:BNNmapping} \cite{FPRRAM, XNOR-RRAM}.

Meanwhile, the number of input neurons on each layer often exceeds the limitation. For example, in case of BNN models proposed by Courbariaux et al. \cite{BNN_bengio}, the minimum length of the input vector except the first layer is 2048 for MNIST dataset and 1152(=3x3x128) for CIFAR-10 dataset. Then, each crossbar array can deal with only part of the input vector and hence needs to process multi-bit partial sums instead of 1-bit output neuron values. 

The partial sums require high resolution for accurate representation, so scalable RRAM-based BNN accelerators require high-resolution ADCs for the partial sums. For example, when the number of rows is 128 for each RRAM array, each array requires 8-bit ADCs to calculate partial sums as shown in Fig. \ref{fig:architecture_prev}(a), and they reduce the efficiency of the accelerator significantly. To handle the issue, BCNN-RRAM \cite{BCNN-RRAM} and XNOR-RRAM \cite{XNOR-RRAM} quantized the partial sum and reduced the resolution of the ADC to 4 bits and 3 bits, respectively. The accelerator architectures of BCNN-RRAM \cite{BCNN-RRAM} and XNOR-RRAM \cite{XNOR-RRAM} are shown in Fig. \ref{fig:architecture_prev}(b) and \ref{fig:architecture_prev}(c), respectively. In BCNN-RRAM \cite{BCNN-RRAM}, 4-bit ADCs still consumed most of total chip power and specific algorithm for quantization and effects of quantization on the inference accuracy were not fully explained. The XNOR-RRAM \cite{XNOR-RRAM} further reduced the resolution of partial sum to 3 bits by using an non-linear quantization. As a trade-off, the XNOR-RRAM required extra circuit components such as look-up tables to handle the non-linear quantization, which increase the area and power consumption. The XNOR-RRAM also used multi-level SA instead of ADC to reduce the power and area overhead but design principle of the multi-level SA is similar to flash ADC, so the overhead still exists.
Thus, a new approach is needed to minimize the overhead of ADC to enable scalable RRAM-based BNN accelerator for the large-scale neural network.

\section{BNN-Hardware Co-design} \label{section:co-design}
In this section, we propose a neural network-hardware co-design method to implement scalable RRAM-based accelerators with 1-bit resolution analog-digital interfaces. In the proposed method, we first reconstruct BNN models considering the hardware constraint of the given RRAM arrays. Then, we set proper parameters for the reconstructed models by manually adjusting parameters of the baseline BNNs. The parameters can be also decided by a neural network training. The potential loss in accuracy from the network reconstruction is recovered through retraining process.

\subsection{Network Reconstruction} \label{subsection:reconstruction}
\begin{figure}[tb]
\begin{center}
\includegraphics[width=0.95\linewidth]{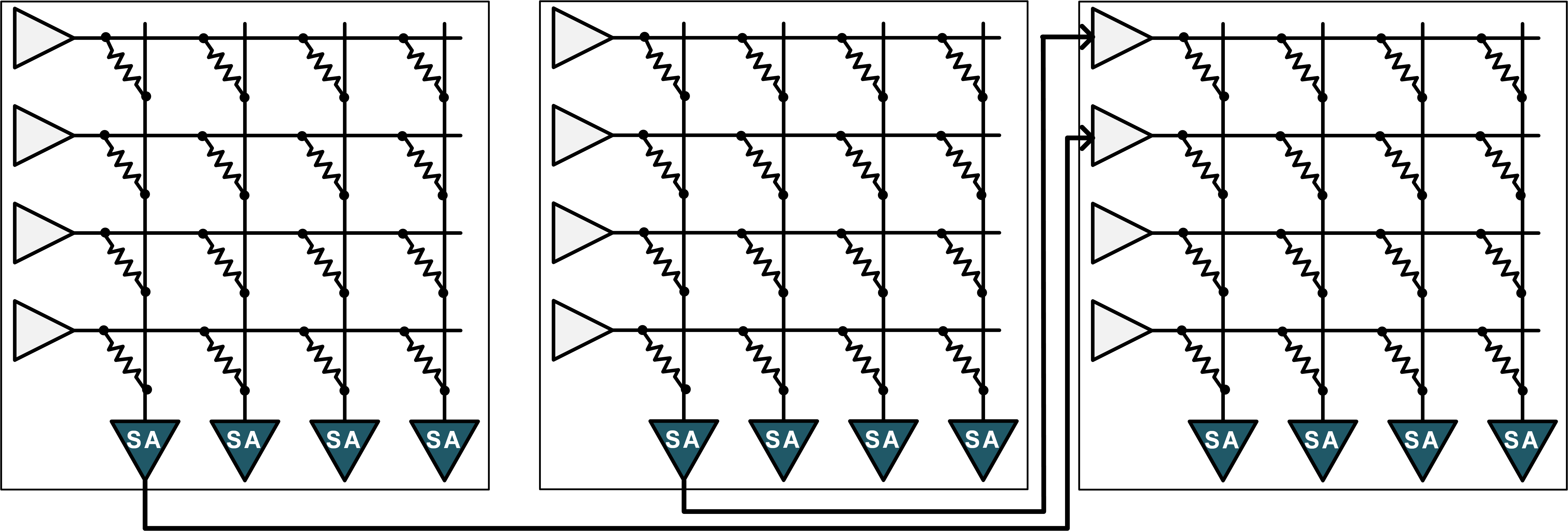}
\caption{Proposed scalable RRAM-based architecture for BNNs}
\label{fig:proposed_architecture}
\end{center}
\end{figure}

For the neural network-hardware co-design, we adopt the input splitting methodology proposed in \cite{ISLPED2018}. Overall architecture of the proposed design is shown in Fig. \ref{fig:proposed_architecture}. The main difference from previous works is that the proposed architecture maintains the 1-bit SAs and word-line drivers while handling the large-scale neural network which needs multiple arrays. In other words, it does not compute the partial sums on each array. Instead, each array produces 1-bit activation values. As a result, overhead of analog-digital interfaces in previous works can be eliminated. 

To successfully run a BNN model on this architecture, the BNN model needs to be modified to fit on the hardware.
Input splitting methodology is used to reconstruct a large BNN layer to a network of smaller layers, which makes each small layer to produce its own 1-bit output in a RRAM array. This makes sure that high-resolution partial sum is not needed any more.

\begin{figure}[tb]
\begin{center}
\includegraphics[width=0.85\linewidth]{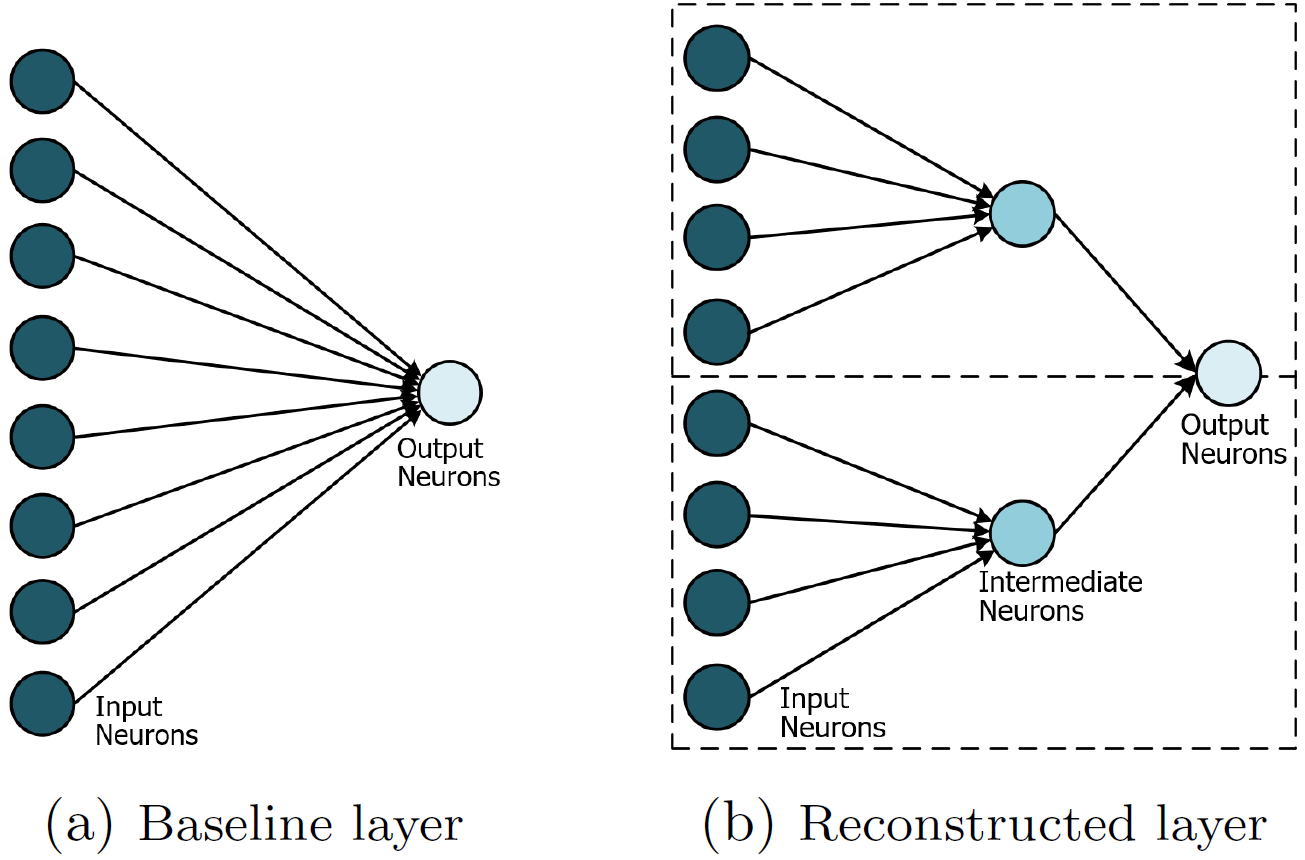}
\caption{An example of reconstructing a BNN layer with two split blocks}
\label{fig:input_splitting}
\end{center}
\end{figure}

The BNN-reconstruction method used in this work consists of the following steps (Fig. \ref{fig:input_splitting}): First, we evenly split input neurons of a BNN layer into several blocks so that the number of inputs per block is smaller than or equal to the number of rows in a crossbar array. Next, we assign intermediate neurons between the input neurons and the output neurons. Because neuron values of the BNN are computed by binarization of weighted-sums of input neurons and weights in the inference phase \cite{FINN}, each block can compute both weighted-sum and binarization using an 1-bit SA. Finally, we connect all intermediate neurons to a corresponding output neuron. To maintain the network complexity of the reconstructed BNN similar to that of baseline BNN, we fix the values of synaptic weights between intermediate neurons and output neurons. We also fix the thresholds of output neurons for binarization. As we use a BNN model \cite{BNN_bengio} that uses (-1, 1) as weight and neuron values in this experiment, we fix the weight values between intermediate and output neurons to 1 and thresholds of output neurons to 0 during the reconstruction. Note that the parameters can be also decided by a neural network training. Fig. \ref{fig:input_splitting} shows an example of reconstructing a BNN layer with two split blocks.

\subsection{Parameter Mapping} \label{subsection:param_mapping}
The reconstructed BNN requires proper parameters to perform inference, and the easiest way to provide parameters for the reconstructed BNN is mapping the parameters of the baseline BNN to the reconstructed network with some adjustments. Because the values of the weights that connect intermediate and output neurons and the thresholds of output neurons are fixed during input splitting, only the weights connecting input and intermediate neurons and thresholds of intermediate neurons need to be determined.
In fact, as the input neurons are connected to intermediate neurons instead of output neurons in the reconstructed BNN, the weights of the baseline BNN can be directly mapped to the weights of the reconstructed BNN.  On the other hand, the thresholds of intermediate neuron binarizations need to be adjusted from the values of the baseline BNN as the number of input neurons becomes different.

Thresholds of neuron binarization depend on the parameters, the neuron biases ($b$) and the batch normalization parameters ($\mu$, $\sigma$, $\gamma$, $\beta$), which shift and scale the weighted-sum results in the training phase \cite{FINN}. 
The shifting-scaling operation of a weighted-sum ($x$) is described as follows,
\begin{equation}
    y = \gamma\frac{(x+b)-\mu}{\sqrt{\sigma^2+\epsilon}}+\beta
    \label{eq:BNorm}
\end{equation}
In the equation, $\epsilon$ is a constant for numerical stability and is typically very small. 

Meanwhile, after input splitting, the sum of shifting-scaling results from split blocks should be equal to results from Eq. \ref{eq:BNorm}. Therefore, for a BNN layer which is split into $n$ blocks, the scaling factor should be the same as that of the baseline layer, and the shifting factor should be $1/n$ that of the baseline layer. Hence, when the parameters of the baseline BNN are mapped to the reconstructed BNN, the $b$, $\mu$, $\sigma$, and $\beta$ are scaled by $1/n$ whereas $\gamma$ remains the same. Then, thresholds of intermediate neurons can be calculated with these parameters with a method proposed by \cite{FINN}.

\subsection{Retraining} \label{subsection:retraining}
Although direct mapping of the weight from the baseline BNN to the reconstructed BNN is possible as shown in the previous section, structural differences exist between the baseline and the reconstructed BNN. In our experiments, it was observed that the parameter mapping could produce reasonable accuracy for a relatively simple case like MNIST dataset, but the accuracy of the reconstructed BNN was significantly degraded when a more complex case such as CIFAR-10 dataset is tested. To recover the accuracy loss, we extensively used the retraining method. As shown in the section \ref{section:Experiments}, the accuracy loss can be substantially recovered by retraining.

For the retraining, we reused the parameter initialization and training conditions that were used for the baseline BNN, which reduced the burden in retraining preparation. The training conditions include image pre-processing, optimizer, and hyper-parameters like learning rate, mini-batch size, dropout ratio, and weight decay. Because the weights which connect the intermediate and output neurons and the thresholds of output neurons were fixed to constant values during input splitting, these parameters were not retrained.

\begin{table}[tb]
\footnotesize
\begin{minipage}{8cm}
\caption{The number of split blocks for the MLP on MNIST} 
\label{MNISTsplit}
\begin{tabular}{M{1.2cm} M{1.8cm} M{1.2cm} M{1.2cm} M{1.2cm}} 
\toprule
\multirow{2}{*}{\hspace{-0.2cm}\begin{tabular}{c}\\[-7pt]\textbf{\# Layer}\end{tabular}} & \multirow{2}{*}{\hspace{-0.2cm}\begin{tabular}{c}\\[-7pt]\textbf{Input count} \\\textbf{per output}\end{tabular}} & \multicolumn{3}{c}{\textbf{Input count per array}}\\ \cmidrule{3-5}
 & & \textbf{512} & \textbf{256} & \textbf{128} \\ \midrule
 1 & 784 & - & - & - \\ 
 2 & 2048 & 4 & 8 & 16 \\
 3 & 2048 & 4 & 8 & 16 \\
 4 & 2048 & - & - & - \\ 
\bottomrule
\end{tabular}
\end{minipage}
\end{table}

\begin{table}[tb]
\footnotesize
\begin{minipage}{8cm}
\caption{The number of split blocks for the CNN on CIFAR-10} 
\label{CIFARsplit}
\begin{tabular}{M{1.2cm} M{1.8cm} M{1.2cm} M{1.2cm} M{1.2cm}} 
\toprule
\multirow{2}{*}{\hspace{-0.2cm}\begin{tabular}{c}\\[-7pt]\textbf{\# Layer}\end{tabular}} & \multirow{2}{*}{\hspace{-0.2cm}\begin{tabular}{c}\\[-7pt]\textbf{Input count} \\\textbf{per output}\end{tabular}} & \multicolumn{3}{c}{\textbf{Input count per array}}\\ \cmidrule{3-5}
 & & \textbf{512} & \textbf{256} & \textbf{128} \\ \midrule
 1 & 3x3x3 & - & - & - \\ 
 2 & 3x3x128 & 3 & 6 & 9 \\
 3 & 3x3x128 & 3 & 6 & 9 \\
 4 & 3x3x256 & 6 & 9 & 18 \\
 5 & 3x3x256 & 6 & 9 & 18 \\
 6 & 3x3x512 & 9 & 18 & 36 \\
 7 & 8192 & 16 & 32 & 64 \\
 8 & 1024 & 2 & 4 & 8 \\
 9 & 1024 & - & - & - \\
\bottomrule
\end{tabular}
\end{minipage}
\end{table}

\begin{figure*}[ht]
\begin{center}
\includegraphics[width=\linewidth]{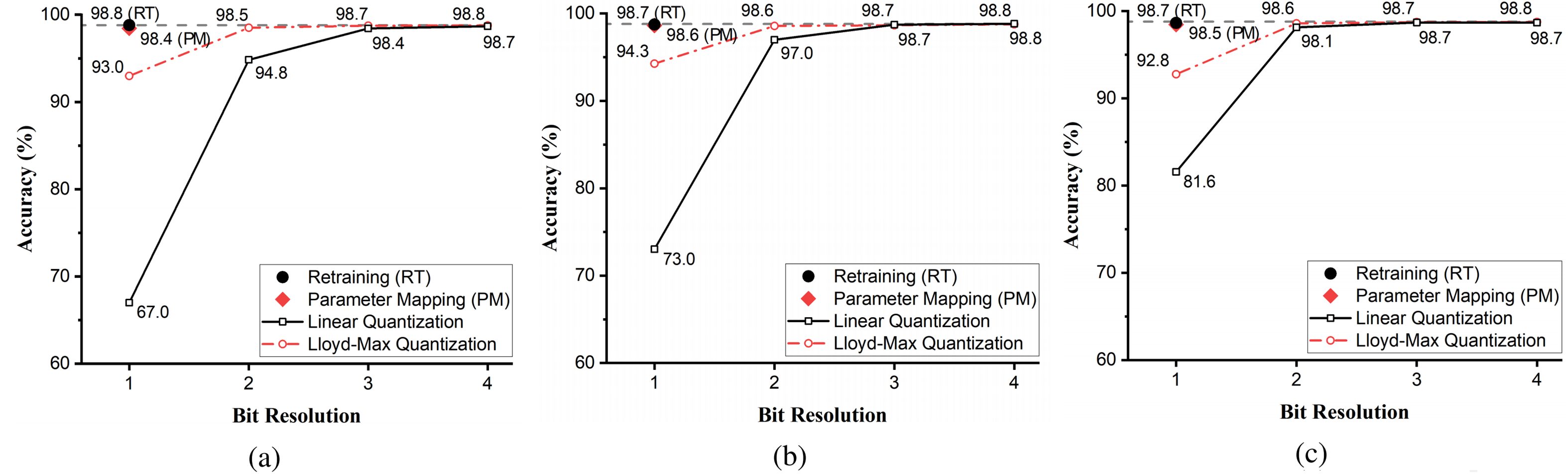}
\caption{Inference accuracy of the MLP on MNIST dataset when input count per array is (a) 512, (b) 256, and (c) 128. (Dashed line: Inference accuracy of the baseline BNN, 98.8\%)}
\vspace{-4mm}
\label{fig:MNISTsplit}
\end{center}
\end{figure*}

\begin{figure*}[ht]
\begin{center}
\includegraphics[width=\linewidth]{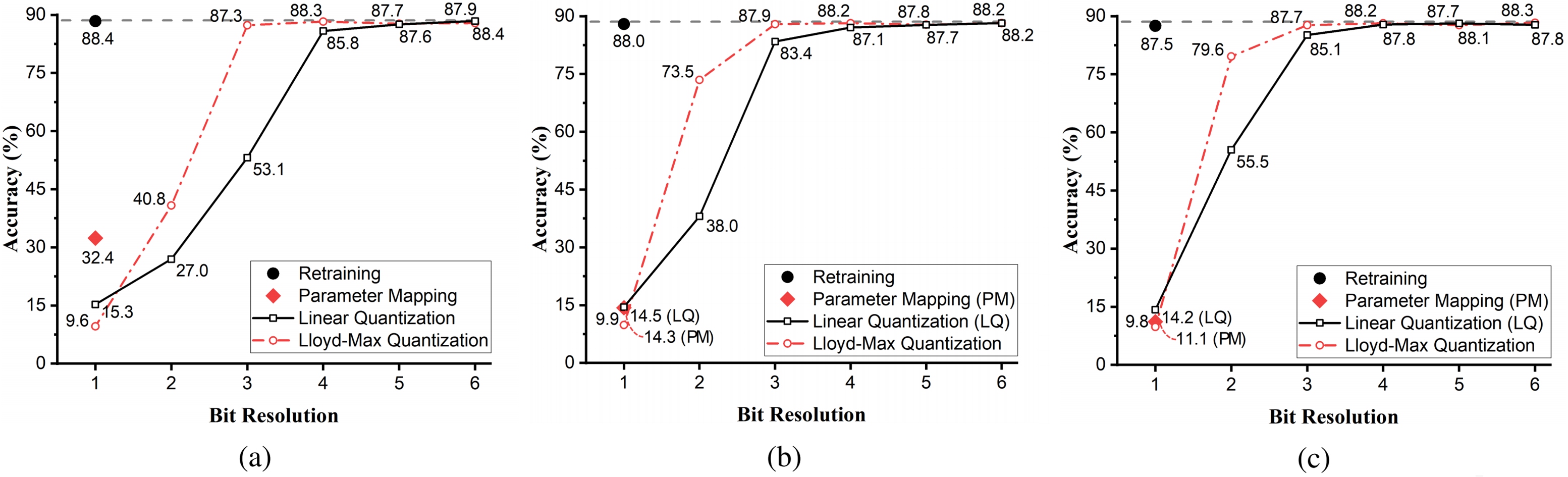}
\caption{Inference accuracy of the CNN on CIFAR-10 dataset when input count per array is (a) 512, (b) 256, and (c) 128. (Dashed line: Inference accuracy of the baseline BNN, 88.6\%)}
\vspace{-5mm}
\label{fig:CIFAR10split}
\end{center}
\end{figure*}

\section{Experimental Results} \label{section:Experiments}
\subsection{Accuracy Analysis}
\textbf{Setup:} 
We used the BNN models proposed in \cite{BNN_bengio} as the baseline BNNs, for which (-1, 1) was used as 1-bit weight and neuron values. The network structure and learning conditions of a multi-layer perceptron (MLP) for MNIST are the same as MLP implemented on Torch7 in \cite{BNN_bengio}, and the structure and learning conditions of a Convolutional Neural Network (CNN) for CIFAR-10 are the same as CNN implemented on Theano in \cite{BNN_bengio}. 

In typical DNN training, shift-based batch normalization is often used to eliminate the multiplication operations in batch normalization during the training phase \cite{BNN_bengio}. In this experiment, general batch normalization \cite{BNorm} was used because the shift-based batch normalization does not contribute much to the speed improvement in BNN training with GPUs and does not affect the speed of inference operation, as the batch normalization parameters are converted into threshold levels for the binarization during the inference \cite{FINN}.

The baseline BNNs are reconstructed by applying input splitting as described in Section \ref{subsection:reconstruction}. Three different RRAM array cases (512, 256, and 128 input neurons per crossbar array) were tried. Note that the first layer of the BNN that processes the input image and the last layer of the BNN that computes the score of each class are excluded from the input splitting. The excluded layers occupy 
only 0.6\% of the total operations in the baseline CNN for CIFAR-10. We summarized the number of blocks per BNN layer after input splitting on Table \ref{MNISTsplit} and Table \ref{CIFARsplit}. Then, we applied both parameter mapping (Section \ref{subsection:param_mapping}) and retraining (Section \ref{subsection:retraining}) to the reconstructed BNNs to determine the synaptic weights.

The results based on the proposed input splitting were compared with the results of previous works, which used quantized partial sums to reduce ADC overhead; one with linear quantization, and the other with non-linear quantization with Lloyd-Max algorithm proposed in XNOR-RRAM \cite{XNOR-RRAM}. Because the Lloyd-Max algorithm requires probability density function (PDF) of the partial sum, we used kernel density estimation with gaussian kernel to estimate PDF, and we set the bandwidth of the estimation ($h$) to $1.06\hat{\sigma} n^{-1/5}$ ($\hat{\sigma}$: the standard deviation for samples, $n$: the number of samples) to minimize the mean integrated squared error \cite{h_KDE}. For fare comparison, we did not quantize partial sum of first and last layers.

\textbf{Results:} 
Fig. \ref{fig:MNISTsplit} and \ref{fig:CIFAR10split} show the accuracy results of BNN models on MNIST and CIFAR-10 dataset, respectively. For all cases, the reconstructed BNNs with retraining showed the inference accuracy close to the accuracy of the baseline BNNs. This result shows that the proposed method can successfully reconstruct BNNs for RRAM-based BNN accelerators with 1-bit analog-digital interfaces as shown in Fig. \ref{fig:proposed_architecture}. In contrast, the inference accuracy of previous BNN models with partial sum quantization depended on both the resolution of quantization and the array size, because the range of partial sum depended on the size of crossbar array. As expected, the partial sum quantization required higher resolution for larger crossbar array. In addition, the higher resolution was required for the partial sums for more complex BNN model.

For the MLP on MNIST dataset, we could keep differences of the inference accuracy between the reconstructed BNN and the baseline BNN within 0.5\% with parameter mapping only, and retraining further reduced the differences. However, linear quantization and Lloyd-Max quantization on partial sum required 3-bit and 2-bit resolution, respectively to keep the accuracy drop within 0.5\%. Both methods experienced significant degradation in the inference accuracy when quantization resolution was reduced to 1 bit. 

For the CNN on CIFAR-10 dataset, relatively large accuracy loss occurred when parameter mapping alone was applied without retraining. The retraining substantially recovered the accuracy loss as shown in Fig. \ref{fig:CIFAR10split}. When the retraining was applied, the differences of inference accuracy between the reconstructed BNN and the baseline BNN were within around 1\% for the worst case. In contrast, partial sum computation with linear qunatization and Lloyd-Max quantization required 6-bit and 4-bit resolution, respectively to keep the accuracy drop around 1\% for the array size 512.

\subsection{Power Analysis}

\textbf{Setup:} We adopted a power estimation model used in BCNN-RRAM \cite{BCNN-RRAM} to analyze the effect of proposed input splitting on power consumption of RRAM-based BNN accelerators. In the power analysis, we focused on RRAM arrays and ADCs, because more than 90\% of processing power is consumed in RRAM arrays and ADCs~\cite{BCNN-RRAM}. 

The system clock and technology node were assumed to be 100MHz and 45nm, respectively. We used 128x128 RRAM arrays with 1T1R RRAM devices to prevent sneak path problem and 4-bit flash ADCs as the baseline \cite{BCNN-RRAM}. In addition, to scale the power of ADCs to the target resolutions, we applied the following equation~\cite{ADCscale},
\begin{equation}
    P_\text{flash} \approx \alpha_2 (2^N-1)
    \label{eq:ADCscale}
\end{equation}

We conducted the experiment on three different resolutions of ADCs (4 bit, 3 bit, and 1 bit) to compare the BCNN-RRAM \cite{BCNN-RRAM}, XNOR-RRAM \cite{XNOR-RRAM}, and the proposed work. The XNOR-RRAM used multi-level SAs but the power consumption is similar to that of flash-type ADCs. Thus, the power consumption of the XNOR-RRAM has been estimated with that of the flash-type ADCs. In addition, we did not consider extra circuit components (e.g. look-up tables) required for handling non-linear quantization in the XNOR-RRAM as the circuit details are not known.

\begin{figure}[tb]
\begin{center}
\includegraphics[width=\linewidth]{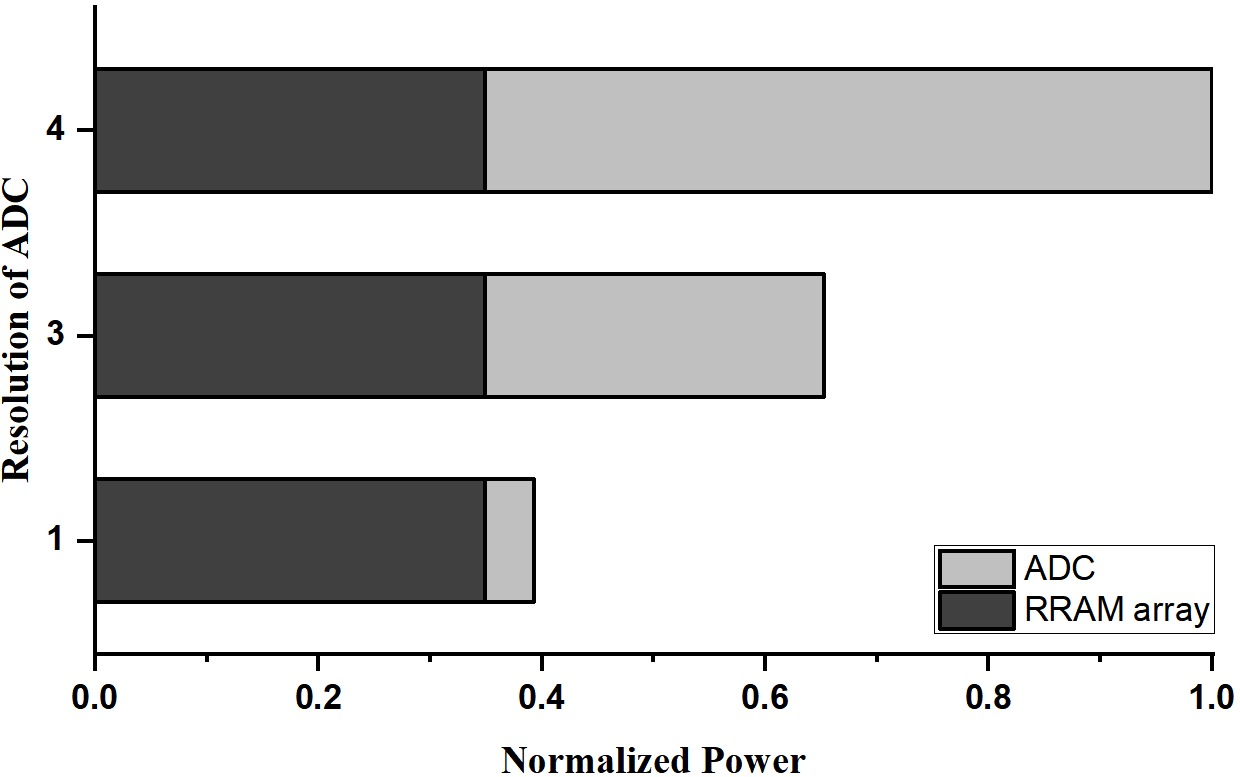}
\caption{Power analysis of RRAM-based BNN accelerators with different resolution of ADCs}
\label{fig:Power}
\end{center}
\end{figure}

\textbf{Results:} Fig. \ref{fig:Power} shows the power analysis result of RRAM-based BNN accelerators with different resolution ADCs. Because the power of the flash ADC decreases exponentially as the bit resolution decreases \cite{ADCscale}, the proposed BNN-hardware co-design methodology can save power consumption in analog-to-digital interfaces compared to the BCNN-RRAM (with 4-bit ADCs) \cite{BCNN-RRAM} and the XNOR-RRAM (with 3-bit ADCs) \cite{XNOR-RRAM} by 93.3\% and 85.7\%, respectively by using 1-bit ADCs. Overall, the proposed method saves power by 60.7\% and 39.9\% compared to the BCNN-RRAM and the XNOR-RRAM, respectively.

\section{Conclusion}
In this paper, we introduced a neural network-hardware co-design method for scalable RRAM-based BNN accelerators. Based on input splitting, parameter mapping, and retraining procedure, the reconstructed BNN allows each RRAM array to have 1-bit analog-digital interface even for large-scale neural network with $\sim1$\% accuracy loss for CIFAR-10 dataset. By using the 1-bit output interface, the proposed design consumes 40\% less power than the previous work. Since maintaining 1-bit interface preserves many promising features of BNN mapped on RRAM array, we believe that the proposed RRAM-based BNN architecture is more scalable, power/area efficient than previous approaches for large-scale neural network implementation.  

\section*{Acknowledgment}
This work was in part supported by Samsung Electronics Co., Ltd. and the Technology Innovation Program (10067764) funded by the Ministry of Trade, Industry \& Energy(MOTIE, Korea). It was also in part supported by the MSIT(Ministry of Science and ICT), Korea, under the ICT Consilience Creative program(IITP-2018-2011-1-00783) supervised by the IITP(Institute for Information \& communications Technology Promotion).


\end{document}